%% file: main.tex
\documentclass[10pt,twocolumn,letterpaper]{article}

\usepackage{cvpr}              
\usepackage[table]{xcolor}

\input{preamble}
\usepackage[acronym]{glossaries}
\usepackage{adjustbox}   
	\usepackage{makecell}  
	\usepackage{float}
	\usepackage{wrapfig}
	\usepackage{booktabs}
	\usepackage{amsmath}
	\usepackage{multirow}
	\DeclareMathOperator*{\argmin}{arg\,min}
\definecolor{cvprblue}{rgb}{0.21,0.49,0.74}
\usepackage[pagebackref,breaklinks,colorlinks,allcolors=cvprblue]{hyperref}

\def\paperID{8} 
\def\confName{CVPR}
\def\confYear{2026}
\newacronym{ppg}{PPG}{Photoplethysmography}
\newacronym{hr}{HR}{Heart Rate}
\newcommand{\modelname}{\textit{PhysioLatent }}
\newacronym{hrv}{HRV}{Heart Rate Variability }
\newacronym{3dvae}{3D VAE}{3D Variational Autoencoder}
\newacronym{adaln}{AdaLN}{Adaptive Layer Normalizations}
\newacronym{film}{FiLM}{Feature-wise Linear Modulation}
\newacronym{mse}{MSE}{Mean Squared Error}
\newacronym{mae}{MAE}{Mean Absolute Error}
\newacronym{rmse}{RMSE}{Root Mean Squared Error}
\newacronym{mape}{MAPE}{Mean Absolute Percentage Error}
\newacronym{rppg}{rPPG}{remote Photoplethysmography}
\newcommand{\green}{\cellcolor{green!20}}

\renewcommand{\red}{\cellcolor{red!20}}
\title{Editing Physiological Signals in Videos Using Latent Representations}

\author{
	Tianwen Zhou\\
	University College London\\
	{\tt\small tianwenzhou0521@gmail.com}
	\and
	Akshay Paruchuri\\
	UNC Chapel Hill\\
	{\tt\small akshay@cs.unc.edu}
	\and
	Josef Spjut\\
	NVIDIA\\
	{\tt\small josef.spjut@gmail.com}
	\and
	Kaan Ak\c{s}it\\
	University College London\\
	{\tt\small kaanaksit@kaanaksit.com}
}
\begin{document}
\maketitle
\input{sec/0_abstract}    
\input{sec/1_intro}

\input{sec/2_relatedwork}
\input{sec/3_method}
\input{sec/4_experiment}

\input{sec/5_discussion}
{
    \small
    \bibliographystyle{ieeenat_fullname}
    \bibliography{main}
}

\input{sec/X_suppl}

\end{document}

%% file: preamble.tex
\newcommand{\red}[1]{{\color{red}#1}}



%% file: sec/0_abstract.tex
\begin{abstract}
Camera-based physiological signal estimation provides a non-contact and convenient means to monitor \gls{hr}. However, the presence of physiological signals in facial videos raises significant privacy concerns, as they can reveal sensitive personal information related to the health and emotional states of an individual. To address this, we propose a learned framework that edits physiological signals in videos while preserving visual fidelity. First, we encode an input video into a latent space via a pretrained 3D Variational Autoencoder (3D VAE), while a target \gls{hr} prompt is embedded through a frozen text encoder. We fuse them using a set of trainable spatio-temporal layers with \gls{adaln} to capture the strong temporal coherence of \gls{rppg} signals. We apply \gls{film} in the decoder with a fine-tuned output layer to avoid the degradation of physiological signals during reconstruction, enabling accurate modulation in the reconstructed video. Empirical results show that our method preserves visual quality with an average PSNR of 38.96 dB and SSIM of 0.98 on selected datasets, while achieving an average \gls{hr} modulation error of 10.00 bpm MAE and 10.09\% MAPE using a state-of-the-art \gls{rppg} estimator. Our design's controllable \gls{hr} editing is useful for applications such as anonymizing biometric signals in real videos or synthesizing realistic videos with desired vital signs.
\end{abstract}

%% file: sec/1_intro.tex
\section{Introduction}
\label{sec:intro}

Physiological signals such as \gls{hr} serve as valuable indicators of a person's physical and emotional states~\cite{Appelhans2006HRV}. With the rise of camera-based physiological sensing technologies \cite{mcduff2023camera}, these signals can now be extracted from facial videos in a non-contact \cite{Cennini:10} and scalable manner \cite{Verkruysse2008RemotePI}, enabling applications in remote health monitoring and human-computer interaction~\cite{orlosky2021telelife}. However, this growing capability also introduces a significant and under-addressed privacy threat. Unlike visible identity cues, physiological signals are often leaked involuntarily while being invisible in human-centric videos \cite{gupta2023privacypreservingremoteheartrate}. When extracted without consent, they can be misused to infer sensitive health conditions, stress levels, or emotional states, raising concerns over covert surveillance, profiling, and medical discrimination. \textit{Most existing visual privacy methods conceal visible identity cues (e.g., face blurring or synthetic replacement) \cite{inbook, DBLP:journals/corr/abs-1803-11556}. While such approaches may unintentionally distort or even suppress the original PPG signal embedded in subtle skin color changes, they often do so at the cost of introducing or amplifying artificial identity-related visual cues, which may be undesirable in privacy-sensitive scenarios.}

\begin{figure}[h]
	\centering
	\includegraphics[width=0.9\linewidth]{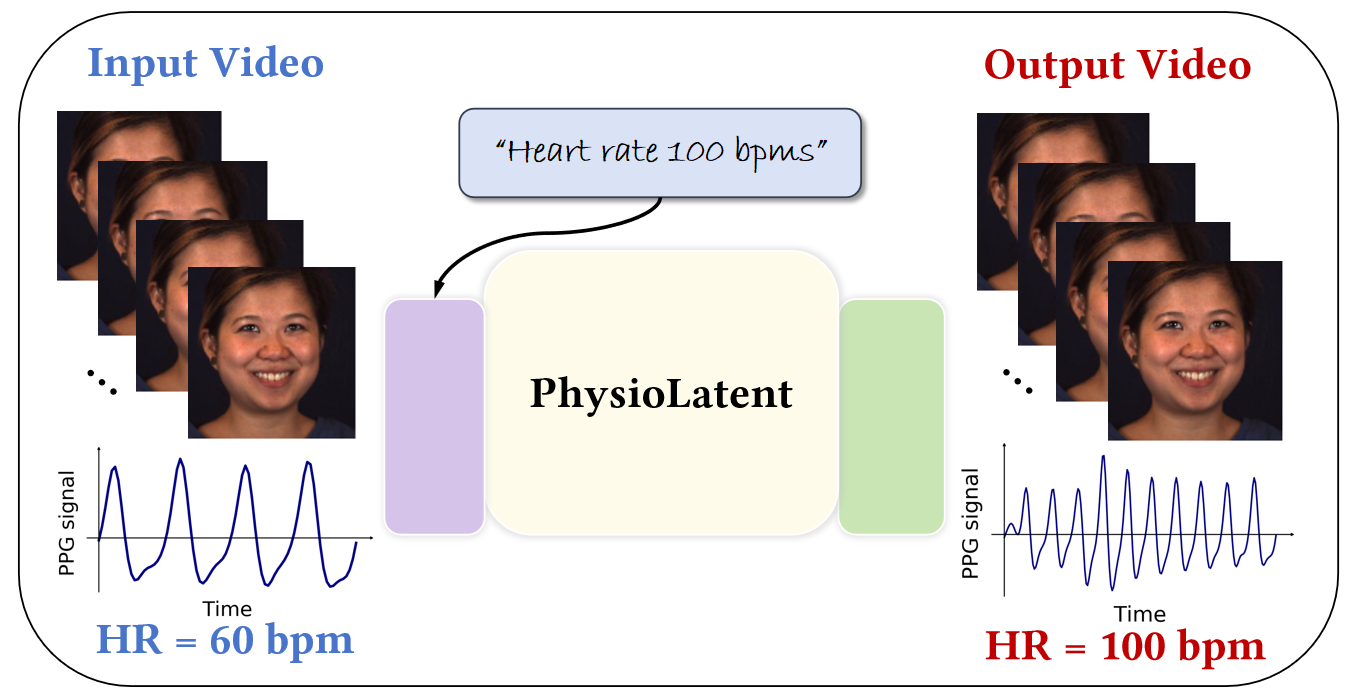}
	\caption{Our learned framework modifies physiological signals in videos by editing them using fused latent representations, maintaining good visual fidelity in the output videos. The input video and \gls{hr} text prompt are encoded and fused to produce the latent representation, which is compatible with foundational generative processes. Source images are from MMSE-HR~\cite{7780743}.}
	\label{fig:teaser}
\end{figure}

We propose a framework for editing physiological signals in videos while preserving visual fidelity. The input video is first encoded into a latent space using a pretrained \gls{3dvae}, and a target \gls{hr} prompt is embedded via a frozen text encoder.  We delibarately choose a 3D VAE latent representation to establish a common latent space that is naturally compatible with foundational generative processes~\cite{Liu2025Survey}, more specifically, Latent Diffusion Models~\cite{rombach2022highresolutionimagesynthesislatent} and Video Diffusion Models~\cite{wan2025, yang2024cogvideox}. This compatibility ensures that our approach can flexibly integrate with, or serve as a post-processing stage for, emerging generative video frameworks. To explicitly exploit the strong temporal coherence of \gls{rppg} signals, we utilize trainable spatio-temporal fusion layers with temporal self-attention module augmented by \gls{adaln}~\cite{peebles2023scalablediffusionmodelstransformers}, allowing precise, long-range temporal conditioning. Moreover, given the subtle amplitude of \gls{rppg} variations, we introduce \gls{film}~\cite{DBLP:journals/corr/abs-1709-07871} in the decoder and fine-tune its output layer, enabling the reconstruction of videos that accurately match the desired physiological modulation. This design ensures both controllable and temporally coherent editing of \gls{hr} signals without sacrificing perceptual quality. Empirical results show that our method preserves visual quality with an average PSNR of 38.96 dB and SSIM of 0.98 on selected datasets, while achieving an average \gls{hr} modulation error of 10.00 bpm MAE and 10.09\% MAPE on selected \gls{rppg} estimator. Our contributions are as follows:

\begin{itemize}
	\item \textbf{Temporal-Aware Latent Fusion.} We design a spatio-temporal latent fusion block with temporal self-attention enhanced by \gls{adaln}. This block helps maintain temporal coherence with the conditioned \gls{rppg} signals.
	
	\item \textbf{Subtle-Signal-Aware Decoder Conditioning.} We address the requirement of preserving small amplitude variations in \gls{rppg} signals by introducing \gls{film} into the \gls{3dvae} decoder and fine-tuning its output layer, enabling accurate physiological modification of target waveforms while maintaining high visual quality.
	
\end{itemize}
Our framework further advances privacy-preserving video sharing, synthetic biometric data generation, and editing of physiological signals in human avatars~\cite{9680677}. Moreover, it can serve as a post-processing stage for generative video models or privacy tools, enabling the synthesis of realistic facial videos with controllable vital signs without degrading fidelity. Our extensive evaluations validate the robustness of the approach, and we will release the codebase publicly upon acceptance.

\begin{figure}
	\centering
	\includegraphics[width=0.6\linewidth]{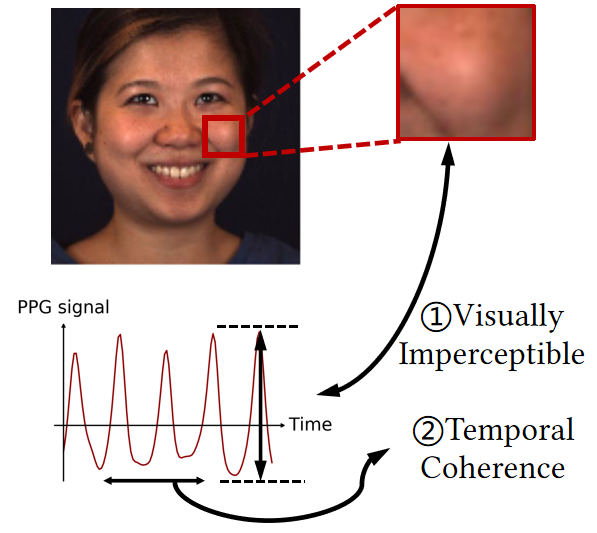}
	\caption{ 
		The two key characteristics of physiological signals in video: temporal coherence and visually imperceptible change. Source images are from MMSE-HR~\cite{7780743}.
	}
	\label{fig:inline}
\end{figure}


%% file: sec/2_relatedwork.tex
\section{Related Works}

We review three relevant categories: (1) Camera-based Physiological Measurement; (2) Manipulating Vital Signs in Videos; and (3) Biometric Privacy Protection. Table. 1 compares our work with the relevant state-of-the-art works in the literature.

\subsection{Camera-based Physiological Measurement}
Monitoring physiological signals, such as \gls{hr} and respiration, is crucial for assessing human emotions and health \cite{Verkruysse2008RemotePI}. \gls{rppg} enables non-contact acquisition of these signals from facial videos. The fundamental mechanism of \gls{rppg} relies on the interaction of light with the multi-layered structure of the skin. Hemoglobin in the dermis partially absorbs light, and a camera reflects and captures the remainder~\cite{Cennini:10}. As blood volume fluctuates with each heartbeat, these changes cause periodic variations in reflected light, which can be captured as subtle color changes in facial regions~\cite{1}. Importantly, such physiological signals possess two characteristics: they exhibit \textbf{strong temporal coherence}, as their variations evolve gradually and rhythmically over time, and they are typically \textbf{visually imperceptible to the human eye}, making them difficult to detect or manipulate directly in pixel space, as visually described in Fig.~\ref{fig:inline}.

\begin{table}[H]
	\centering
	
	\renewcommand{\arraystretch}{1.3} 
	
	\caption{
		Comparison of video-based vital sign manipulation methods. 
		Columns indicate whether a method supports \gls{hr} editing, removing physiological signals, preserving temporal coherency in edited physiological signals, protecting biometric privacy, focusing on motion robustness.
	}
	\label{tab:related} 
	
	\begin{adjustbox}{width=\columnwidth}
		\begin{tabular}{lccccc}
			\hline
			& \textbf{HR Edit} & \makecell[c]{\textbf{Signal}\\ \textbf{Removal}} & \makecell[c]{\textbf{Temporal} \\ \textbf{Coherency}} & \textbf{Privacy} & \makecell[c]{\textbf{Motion} \\ \textbf{Robustness}} \\
			\hline
			\textbf{This work} & \green Yes & \green Yes & \green High & \green Yes & \red No \\
			Chen et al.~\cite{9680677} & \green Yes & \green Yes & \red Low & \green Yes & \red No \\
			Wang et al.~\cite{9878643} & \green Yes & \red No & \green High & \red No & \red No \\
			Paruchuri et al.~\cite{paruchuri2024motion} & \red No & \red No & \green High & \red No & \green Yes \\
			Sun et al.~\cite{9806161} & \green Yes & \green Yes & \red Low & \green Yes & \red No \\
			\hline
		\end{tabular}
	\end{adjustbox}
\end{table}

Poh et al.~\cite{Poh:10} were among the first to demonstrate an accurate and low-cost \gls{rppg} method using a webcam, which combined automatic face tracking with blind source separation to extract \gls{hr}. Since then, various signal extraction pipelines and benchmark tools, such as the \gls{rppg} toolbox \cite{liu2022rppg}, have been developed to evaluate algorithms under standardized settings. Recent surveys by McDuff~\cite{mcduff2023camera} and Xiao et al.~\cite{XIAO2024105608} have reviewed progress in camera-based physiological sensing, covering signal origin, extraction pipelines, and application domains. For completeness, we also briefly review estimation techniques, as they are used for evaluation in our experiments. Conventional \gls{rppg} pipelines typically begin with face detection, followed by Region of Interest (ROI) selection (e.g., forehead or cheeks), temporal averaging, and blind source separation such as PCA~\cite{10.1145/3447755}, POS~\cite{wang2017algorithmic}, and CHROM~\cite{6523142}. More recently, deep learning models including TSCAN~\cite{liu2021multitasktemporalshiftattention}, DeepPhys~\cite{chen2018deepphys}, PhysNet~\cite{DBLP:confbmvcYuLZ19}, and PhysFormer++~\cite{10.1007/s11263-023-01758-1} have been proposed to directly regress \gls{hr} from video using CNNs or transformers. \textit{In our work, we adopt well-established methods~\cite{10.1145/3447755, wang2017algorithmic, 6523142, inproceedings2018, chen2018deepphys, DBLP:confbmvcYuLZ19, 10.1007/s11263-023-01758-1} to validate the accuracy of physiological signal editing in videos.} 

\begin{figure*}[!htbp]
	\centering
	\includegraphics[width=\linewidth]{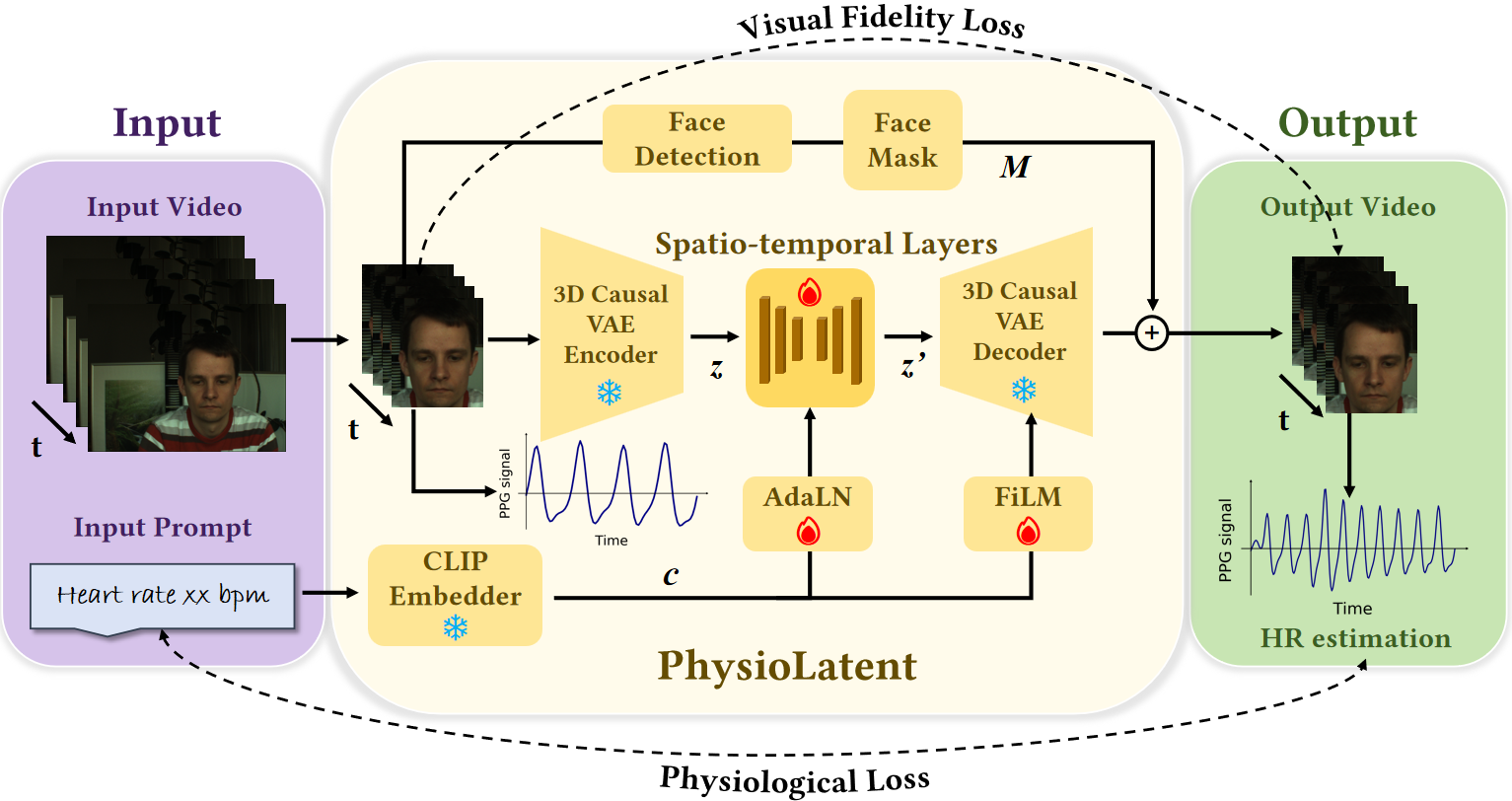}
	\caption{
		Overview of the proposed framework. An input facial video and a \gls{hr} prompt are encoded into latent representations \(z\) and \(c\) using a 3D Causal VAE~\cite{yang2024cogvideox} encoder and a CLIP text embedder, respectively. The fused latent features are processed by spatio-temporal layers with \gls{adaln} to inject temporal coherence and subtle \gls{rppg} variations. The 3D Causal VAE decoder reconstructs the output video with \gls{film} conditioning, which is supervised by visual fidelity and physiological losses to ensure perceptual quality and accurate \gls{hr} modulation. To enhance visual fidelity, we incorporate a face detection module that generates a face mask \(M\), replacing only the facial region of the input video with the decoder output. Source images are from PURE~\cite{stricker2014video}.
	}
	\label{fig:overview}
\end{figure*}

\subsection{Manipulating Vital Signs in Videos}
Recent works have explored ways to manipulate or synthesize such signals, especially for privacy, controllability, and data augmentation. PulseEdit \cite{9680677} introduced a method that perturbs the underlying \gls{rppg} signal in videos—either suppressing it or modifying it toward a target waveform—while preserving the visual appearance. Similarly, Wang et al.~\cite{9878643} proposed a generative adversarial framework that synthesizes realistic face videos conditioned on \gls{rppg} signals, enabling control over the generated physiological patterns. Most recently, Paruchuri et al.~\cite{paruchuri2024motion} proposed a motion augmentation framework that reasonably preserves \gls{hr} while using natural videos (e.g., interviews on YouTube) to inject or modify rigid and non-rigid facial motions, enabling more robust physiological sensing under realistic motion conditions. The physiological signals' successful manipulation requires models to preserve continuity and consistency across time, while ensuring that modifications remain visually unobtrusive. \textit{Unlike prior methods replacing or removing HR on the image without any intermediate representations, our work edits HR directly in latent space, enabling temporally coherent and visually faithful physiological signal manipulation. Usage of 3D VAE latent space reflects a growing interest towards physiologically-conditioned generative modeling.}
\subsection{Biometric Privacy Protection}
Traditional biometric privacy protection algorithms \cite{7192837} focus on de-identifying facial or fingerprint features to prevent identity theft. Recent work \cite{Bhutani2025Privacy} has shown that physiological signals such as \gls{hr} and stress level can also be inferred from facial videos without consent, raising concerns about profiling and emotional inference. PulseEdit \cite{9680677} introduced a perturbation-based way that removes or alters physiological signals in facial videos while preserving visual appearance, thereby reducing privacy leakage. De-id~\cite{Savic_2023_BMVC} proposed a learning-based facial video de-identification method that eliminates facial identity from videos while retaining the embedded \cite{9806161} signals and visual quality. Privacy-Phys \cite{9806161} presented a 3D CNN-based framework for modifying \gls{rppg} signals in facial videos to prevent physiological information leakage. \textit{Our work shares the same privacy-preserving motivation of these studies}.

%% file: sec/3_method.tex
\section{Editing Physiological Signs Using Latent Representations}
We begin by defining our research problem, followed by an overview of our proposed framework for editing and anonymizing \gls{rppg} signals in video clips while preserving visual fidelity.

\subsection{Problem Definition}
Let \( v_t \in \mathbb{R}^{C \times H \times W} \) denote the \( t \)-th frame of a video, where \( C \), \( H \), and \( W \) are the number of color channels, height, and width. Each frame \( v_t \) encodes both visual appearance \( v_t^{\text{vis}} \) and biological signal components \( v_t^{\text{bio}} \), such that:
\begin{equation}
	v_t = v_t^{\text{vis}} + v_t^{\text{bio}}.
\end{equation}
where \( v_t^{\text{vis}} \) denotes the dominant visual appearance component of the frame, including facial identity, texture, and illumination, while \( v_t^{\text{bio}} \) represents the latent physiological signal component manifested as subtle, temporally varying color fluctuations correlated with biological rhythms (e.g., \gls{hr}). 
Given a video \( v \) and a condition vector \( c \in \mathbb{R}^m \) specifying the target \gls{hr}, our goal is to modify the biological component \( v_t^{\text{bio}} \) while preserving the visual component \( v_t^{\text{vis}} \). Formally, we aim to learn a transformation function \( F(\cdot) \) that produces an anonymized video \( \hat{v} = \{\hat{v}_1, \hat{v}_2, \dots, \hat{v}_T\} \) such that
\begin{equation}
	\hat{v}_t = F(v_t, c) = v_t^{\text{vis}} + \hat{v}_t^{\text{bio}}.
\end{equation}
The anonymized biological signal \( \hat{v}_t^{\text{bio}} \) should satisfy these criteria:
\begin{itemize}
	\item \textbf{Privacy Preservation}: The modified signal should reflect the desired \gls{hr} \( HR_d \), i.e., \( HR(\hat{v}_t^{\text{bio}}) \approx HR_d \).
	\item \textbf{Visual Fidelity}: The perceptual quality of the visual content must remain close to the original, i.e., \( \mathcal{F}(v_t^{\text{vis}}, \hat{v}_t^{\text{vis}}) \leq k \), where \( \mathcal{F} \) is a visual similarity measure (e.g., \gls{mse}, LPIPS~\cite{zhang2018unreasonable}).
	\item \textbf{Computational Efficiency}: The transformation function \( F(\cdot) \) should be efficient enough to support real-time applications.
\end{itemize}

To this end, we formulate the following optimization problem:
\begin{equation}
	\tilde{v}_t \leftarrow \argmin_{v_t} \;
	\mathcal{L}\!\big(F(v_t, c), \hat{v}_t\big),
\end{equation}
where the loss function $\mathcal{L}$ will be detailed in Sec.~\ref{sec:loss}. In addition, it is also important to note that \gls{rppg} signals exhibit two key characteristics visually described as in Fig~\ref{fig:inline}: (1) \textbf{strong temporal dependency}, as the periodic rhythm of heartbeat manifests across multiple frames; and (2) \textbf{subtle pixel-level variations}, typically of less than 1\% change in intensity. \textit{Our study reveals that maintaining required qualities in the final videos dictates a dedicated spatio-temporal modeling, see Sec.~\ref{sec:stl} and Sec.~\ref{sec:vilm} for details.}

\subsection{PhysioLatent Model Overview}
We provide an illustration of our overall framework \modelname in Figure~\ref{fig:overview}.
An input video clip is first encoded into a latent representation by a frozen 3D Causal VAE~\cite{yang2024cogvideox} encoder, while a target \gls{hr} prompt is projected via a CLIP text embedder. 
These two embeddings are fused and processed by trainable spatio-temporal layers designed to capture the temporal coherence of \gls{rppg} and preserve spatial fidelity. 
The modified latent code is decoded by a pretrained 3D Causal VAE decoder, where \gls{film}~\cite{DBLP:journals/corr/abs-1709-07871} conditioning injects the desired physiological variation and the output layer is fine-tuned to adapt to subtle \gls{rppg} edits. 
To further preserve visual fidelity, we use an off-the-shelf Haar Cascade~\cite{ViolaJones2001} face detector to localize faces and replace the facial region of the input with the decoded output, keeping the rest of the frame unchanged. 
To guide the transformation, the output is supervised by a visual fidelity loss and a physiological loss that enforces alignment with the desired \gls{hr}. 
This design provides a clear separation between identity-preserving visual content and condition-controlled physiological signals, which will be detailed in the following subsections.

\subsubsection{Latent Space Conditioning}
We firstly construct a \gls{hr} prompt in natural language with a fixed prompt template—e.g., \textit{“Heart rate xx bpm”}, where \( xx \in [60, 120] \)—and encode it using a frozen CLIP~\cite{radford2021learning} text encoder. We choose CLIP embeddings over a one-hot encoding scheme not only because CLIP provides stronger representational capacity, but also because it is more naturally compatible with future generative editing pipelines, as illustrated in Sec.~\ref{sec:application}. The resulting 512-dimensional embedding is projected via a learnable linear layer to match the spatial dimensions of the video latent features. The projected prompt token is concatenated channel-wise with the video latent tensor and passed to the spatio-temporal layers for conditional transformation.

\subsubsection{Spatio-temporal Layers}
\label{sec:stl}
Biologically plausible modification in latent space demands modelling both spatial structure and temporal dynamics (see Fig.~\ref{fig:inline}). Spatial features encode facial geometry and texture, while temporal cues, such as color fluctuations and micro-motions, play a crucial role in \gls{hr} modulation.

\begin{figure}[h]
	\centering
	\includegraphics[width=\linewidth]{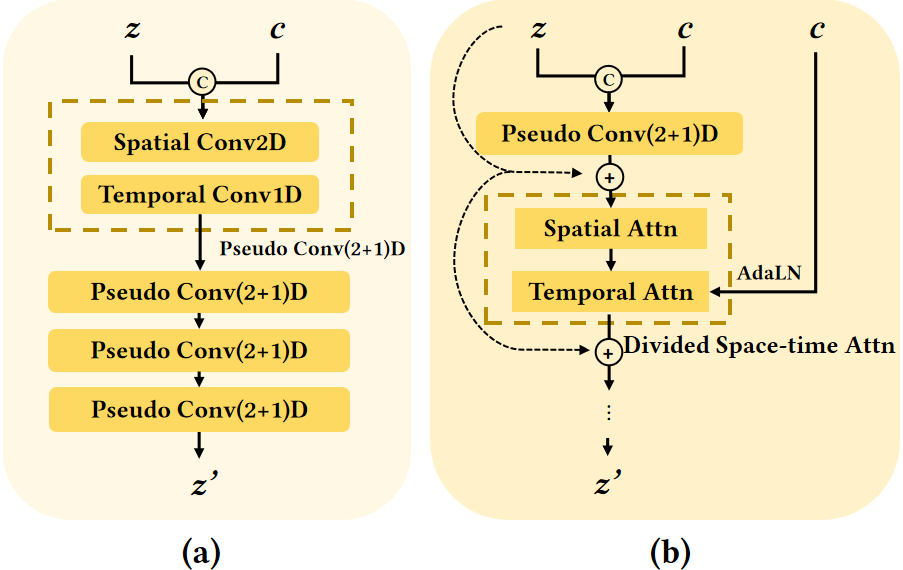}
	\caption{Comparison of proposed spatio-temporal layer settings. (a) A naïve baseline that fuses features using only stacked pseudo (2+1)D convolutions, which captures local patterns but fails to model long-range dependencies. (b) Our improved design that addresses the temporal correlation of \gls{rppg} signals by introducing decomposable space-time self-attention, with \gls{adaln} injecting the desired \gls{hr} signal into the temporal stream for fine-grained modulation. $z$ and $z'$ denote the input and output latent vectors, and $c$ corresponds to the embedded text prompt.}
	\label{fig:settings}
\end{figure}

The model sketched in Fig.~\ref{fig:settings}(a) represents a naïve yet commonly used architecture for spatio-temporal fusion, relying solely on stacked pseudo (2+1)D convolutions~\cite{qiu2017learningspatiotemporalrepresentationpseudo3d}. While it captures local patterns, it lacks the ability to explicitly model long-range dependencies critical to \gls{rppg} signal propagation.
In contrast, setting shown in Fig.~\ref{fig:settings}(b) addresses the \textbf{temporal correlation} requirement of rPPG signals. It incorporates a decomposable space-time self-attention module, allowing the model to learn long-range dependencies  effectively. Moreover, to \textbf{\textbf{inject the desired \gls{hr} signal into the temporal stream}}, we employ \gls{adaln} within the time-attention block, enabling fine-grained modulation of latent features conditioned on the \gls{hr} prompt.

\subsubsection{Visually Imperceptible Learned Modifications}
\label{sec:vilm}
Give the visually imperceptible subtle nature of \gls{rppg} signals (see Fig.~\ref{fig:inline}), we further enhance the decoder's sensitivity to the desired physiological signal by introducing explicit conditioning. We employ a \gls{film}~\cite{DBLP:journals/corr/abs-1709-07871} layer within the 3D Causal VAE decoder. The CLIP-projected \gls{hr} embedding is used to generate scaling and shifting parameters, which modulate the intermediate activations of the decoder. This conditioning mechanism allows the framework to better reconstruct videos that conform to the specified \gls{hr}. Furthermore, we unfreeze the output layer of the decoder and fine-tune it during training. This choice stems from our observation that the pre-trained decoder, while powerful for generic video reconstruction, is not optimized for capturing the subtle \gls{rppg} dynamics. Partial fine-tuning improves adaptability without losing the pretrained knowledge.

\subsubsection{Loss Function}
\label{sec:loss}
The training objective combines both visual and physiological constraints to ensure that the edited video maintains perceptual realism while aligning with the desired \gls{hr}.

\noindent\textbf{Visual Fidelity Loss} (\( \mathcal{L}_{F} \)):  
We adopt a hybrid loss combining pixel-wise \gls{mse} and LPIPS to preserve visual quality:
\begin{equation}
	\mathcal{L}_{F} = \text{MSE}(v, \hat{v}) + \text{LPIPS}(v, \hat{v}).
\end{equation}

\noindent\textbf{Wave Loss} (\( \mathcal{L}_{\text{wave}} \)):  
To enforce temporal coherence and guide the model toward the desired waveform, we compute the negative Pearson correlation between the anonymized \gls{rppg} signal and a target sinusoidal reference with frequency \(f = HR_d / 60\):
\begin{equation}
	\mathcal{L}_{\text{wave}} = 1 -\text{Pearson}\big(rPPG(\hat{v}), \sin(2\pi f t)\big).
\end{equation}
From a morphological perspective, the Pearson correlation emphasizes the similarity of the waveform shape, thereby guiding the anonymized rPPG toward a smooth, periodic morphology while suppressing subject-specific details.

\noindent\textbf{Frequency Loss} (\( \mathcal{L}_{\text{freq}} \)):  
To align the peak frequency \(f_{\text{pred}}\) of the estimated signal with the target frequency, we define:
\begin{equation}
	\mathcal{L}_{\text{freq}} = \big| f - f_{\text{pred}} \big|,
\end{equation}
where \(f_{\text{pred}}\) is obtained via Fast Fourier Transform (FFT)~\cite{cooley1965algorithm}.  
In a curriculum learning manner, \( \mathcal{L}_{\text{freq}} \) is excluded during the first ten epochs, allowing the model to focus on reconstructing coarse visual structures. From epoch ten onward, its contribution is gradually increased via a weight function \(\beta(t)\), as shown in Eq.~\ref{eq:beta}.

\noindent\textbf{Total Loss:}
\begin{equation}
	\mathcal{L}_{\text{total}} 
	= 
	\underbrace{\alpha \mathcal{L}_{\text{wave}} + \beta(t)\mathcal{L}_{\text{freq}}}_{\text{Physiological Loss}}
	+ \lambda \mathcal{L}_{F},
\end{equation}
where \(\alpha\) controls the wave loss, \(\beta(t)\) controls the frequency loss for epoch $t$, and \(\lambda\) balances the fidelity term against the physiological terms. In our experiment, we set the weights as $\alpha = 0.2, \lambda = 1.0$, and use a warm-up schedule for the frequency loss:
\begin{equation}
	\label{eq:beta}
	\beta(t)=
	\begin{cases}
		0, & t \le 10,\\[4pt]
		0.005\,(t-10), & t > 10,\\[4pt]
	\end{cases}
\end{equation}
so that the frequency loss is disabled in the first ten epochs, then linearly ramped from 0 over the subsequent epochs. 

%% file: sec/4_experiment.tex
\section{Evaluation}
We detail our experimental settings used for all evaluations. Then, we provide a detailed analysis in qualitative and quantitative terms. Lastly, we provide benchmark comparison and an ablation study to justify our design choices.
\subsection{Implementation}
We adopt the publicly available 3D Causal VAE~\cite{yang2024cogvideox} model as our video compressor. This model provides a four times downsampling in spatial resolution and an eight times downsampling in the temporal dimension. For text embedding, we utilize the pretrained CLIP~\cite{radford2021learning} model \texttt{ViT-B/32} as the prompt encoder.
Our framework is trained using four NVIDIA RTX 4090 GPUs, each with 24 GB of memory. We use a batch size of 4 and conduct the training for 30 epochs. The training is optimized using the AdamW~\cite{paszke2017automatic} optimizer, and a OneCycle learning rate scheduler~\cite{DBLP:journals/corr/abs-1708-07120} is employed, with base learning rate set to $5 \times 10^{-4}$.
Prior to being fed into the 3D Causal VAE encoder, each input video is temporally chunked into clips of 128 frames and spatially cropped to retain only the facial region. Following the resolution used in the state of the art literature~\cite{liu2022rppg,DBLP:confbmvcYuLZ19}, we resize the resolution of facial region to 72 px by 72 px. We adopt the well-established POS estimator~\cite{wang2017algorithmic}  throughout all stages of our framework, following its widespread use in recent rPPG works~\cite{liu2022rppg}. POS provides stable and consistent pulse estimations compared with alternative estimators~\cite{10.1007/s11263-023-01758-1,chen2018deepphys} as evidenced evaluations in our Table ~\ref{tab:estimators}.

\textbf{Dataset.} 
We adopt multiple publicly available \gls{rppg} datasets to enable both within-dataset and cross-dataset evaluation. 
The PURE dataset~\cite{stricker2014video} consists of 59 facial video recordings captured at a resolution of $640\times480$ pixels and a frame rate of 30 frames per second (fps). All recordings were collected in well-lit indoor environments from a total of 10 subjects, each recorded under six different conditions: steady, talking, slow translation, fast translation, small rotation, and medium rotation. We use an 80/20 split of the dataset for training and testing, respectively.  In addition, we include the UBFC-rPPG dataset~\cite{bobbia2017unsupervised}, which contains 42 videos from 42 subjects recorded under relatively constrained conditions. 
To enhance diversity in subject appearance, we also evaluate on the MMPD dataset~\cite{10340857}, which includes subjects across different skin tone types and a variety of illumination and motion conditions.  
This combination of datasets allows us to assess the robustness of our method under varying recording and camera conditions, subject diversity, and domain shift. Our face detection, cropping, and mask generation follow the standardized implementation in the \gls{rppg} toolbox~\cite{liu2022rppg}, which specifies detector configuration, thresholds, and failure-handling strategies; therefore, we directly adopt this pipeline without further modification.

\begin{figure}[htbp]
	\centering
	\includegraphics[width=\linewidth]{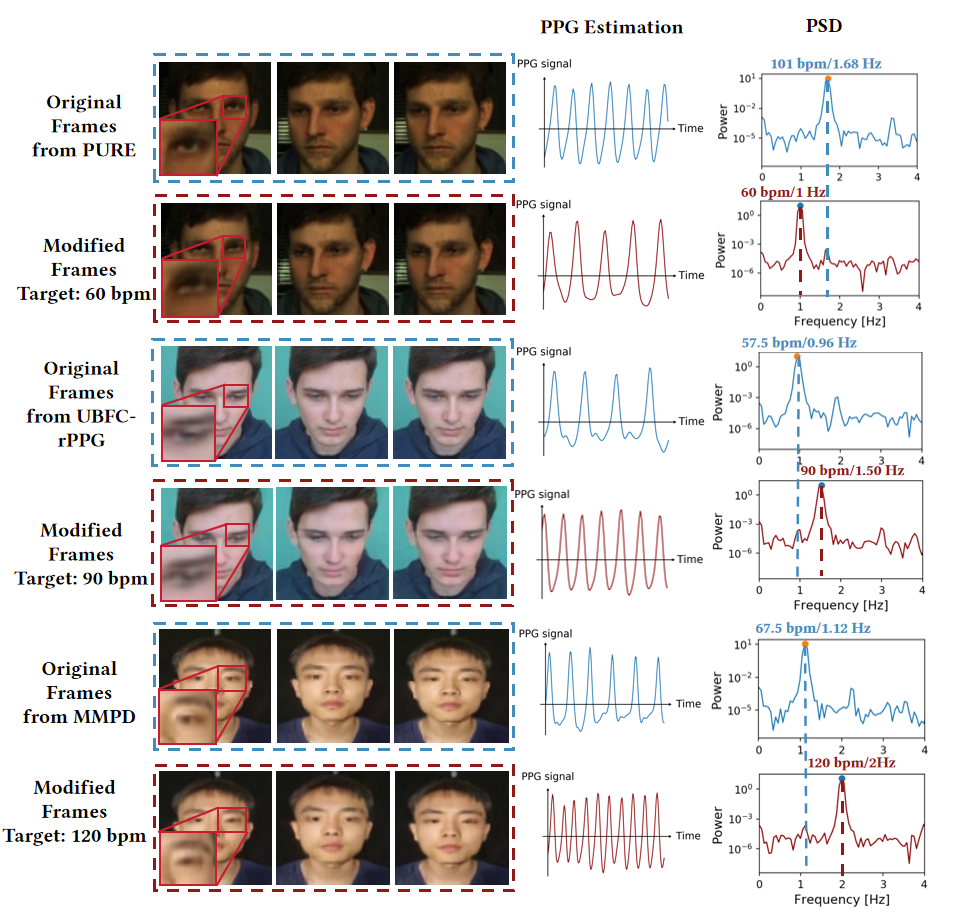}
	\caption{Qualitative comparison results of our proposed framework. We visualize representative novel frames from multiple datasets before and after modification, together with the estimated rPPG signals under the POS estimator and the corresponding Power Spectrum Density (PSD) curves. We target \gls{hr} values commonly encountered in real-life scenarios. In addition, we perform zooming on selected regions of the frames to better reveal fine-grained details. From zoomed-in results, it can be observed that due to the encoding–decoding pipeline of the \gls{3dvae} backbone, local distortions appear in high-frequency areas such as edges and textures. Source images are from MMPD~\cite{10340857}.}
	\label{fig:comparison}
\end{figure}
\begin{table*}[htbp]
	\centering
	\caption{Comparison of video quality and \gls{rppg} estimation accuracy between input videos and outputs produced by our framework across three benchmark datasets. 
		PSNR and SSIM demonstrate that the visual fidelity of videos is well preserved. 
		For physiological signal analysis, the \gls{hr}s are estimated under the POS estimator. We report MAE and MAPE as mean $\pm$ standard deviation with respect to the selected target \gls{hr}s. 
		Results highlight that the input signals exhibit large errors from the target \gls{hr}, while our outputs consistently achieve substantially lower MAE/MAPE, indicating the success of the \gls{hr} modification towards the target.}
	\label{tab:metrics}
	\begin{adjustbox}{max width=0.9\textwidth}
		\begin{tabular}{c c c c|cc|cc}
			\toprule
			\multirow{2}{*}{\textbf{Dataset}} & \multirow{2}{*}{\textbf{Target HR(bpm)}} 
			& \multirow{2}{*}{\textbf{PSNR(dB)} $\uparrow$} & \multirow{2}{*}{\textbf{SSIM} $\uparrow$} 
			& \multicolumn{2}{c|}{\textbf{Input \gls{rppg} Accuracy}} 
			& \multicolumn{2}{c}{\textbf{Output \gls{rppg} Accuracy}} \\
			\cmidrule(lr){5-6} \cmidrule(lr){7-8}
			& & & 
			& MAE(bpm) $\downarrow$ & MAPE(\%) $\downarrow$ 
			& MAE(bpm) $\downarrow$ & MAPE(\%) $\downarrow$ \\
			\midrule
			\multirow{5}{*}{\makecell{PURE\\\cite{stricker2014video}}} 
			& 60  & 39.04 & 0.98 
			& 38.95 ($\pm$13.8) & 31.10 ($\pm$12.9) 
			& 9.22 ($\pm$4.4) & 8.20 ($\pm$4.1) \\
			& 80  & 39.02 & 0.98 
			& 41.80 ($\pm$13.5) & 35.88 ($\pm$12.7) 
			& 9.98 ($\pm$4.0) & 10.55 ($\pm$3.6) \\
			& 100 & 38.85 & 0.98 
			& 44.67 ($\pm$12.9) & 38.92 ($\pm$12.1) 
			& 10.41 ($\pm$4.1) & 10.29 ($\pm$3.9) \\
			& 120 & 38.94 & 0.98 
			& 50.63 ($\pm$14.2) & 42.20 ($\pm$13.7) 
			& 10.36 ($\pm$4.5) & 11.34 ($\pm$4.0) \\
			& \textbf{Avg} & \textbf{38.96} & \textbf{0.98} 
			& \textbf{44.01} & \textbf{37.02} 
			& \textbf{10.00} & \textbf{10.09} \\
			\midrule
			\multirow{5}{*}{\makecell{UBFC-rPPG\\~\cite{bobbia2017unsupervised}}} 
			& 60  & 39.86 & 0.98 
			& 31.91 ($\pm$7.6) & 33.82 ($\pm$7.3) 
			& 11.05 ($\pm$4.0) & 10.08 ($\pm$3.6) \\
			& 80  & 39.98 & 0.98 
			& 27.17 ($\pm$9.0) & 24.01 ($\pm$8.7) 
			& 12.28 ($\pm$4.5) & 11.05 ($\pm$3.9) \\
			& 100 & 40.11 & 0.98 
			& 23.65 ($\pm$8.9) & 19.02 ($\pm$8.3) 
			& 9.82 ($\pm$3.0) & 10.98 ($\pm$2.8) \\
			& 120 & 40.04 & 0.98 
			& 24.40 ($\pm$7.9) & 20.34 ($\pm$7.5) 
			& 11.18 ($\pm$4.1) & 10.15 ($\pm$3.7) \\
			& \textbf{Avg} & \textbf{40.09} & \textbf{0.98} 
			& \textbf{26.77} & \textbf{24.30} 
			& \textbf{11.08} & \textbf{10.57} \\
			\midrule
			\multirow{5}{*}{\makecell{MMPD\\~\cite{10340857}}} 
			& 60  & 36.95 & 0.94 
			& 47.57 ($\pm$16.2) & 33.65 ($\pm$15.5) 
			& 10.62 ($\pm$4.3) & 7.88 ($\pm$3.5) \\
			& 80  & 37.56 & 0.95 
			& 42.85 ($\pm$15.8) & 35.45 ($\pm$15.0) 
			& 10.38 ($\pm$4.1) & 9.82 ($\pm$3.6) \\
			& 100 & 37.60 & 0.94 
			& 46.85 ($\pm$16.9) & 38.82 ($\pm$16.0) 
			& 7.62 ($\pm$3.4) & 6.70 ($\pm$2.9) \\
			& 120 & 37.87 & 0.95 
			& 41.03 ($\pm$16.5) & 34.19 ($\pm$15.9) 
			& 10.75 ($\pm$4.4) & 7.96 ($\pm$3.6) \\
			& \textbf{Avg} & \textbf{37.50} & \textbf{0.95} 
			& \textbf{44.58} & \textbf{35.52} 
			& \textbf{9.84} & \textbf{8.09} \\
			\bottomrule
		\end{tabular}
	\end{adjustbox}
\end{table*}

\textbf{Evaluation Metrics.} 
Our assessments measure how well the modified videos align with the desired objectives. For visual fidelity, we employ PSNR and SSIM. To evaluate the effectiveness of \gls{hr} modulation, we employ several established \gls{rppg} estimation algorithms. Specifically, we utilize CHROM \cite{6523142}, PCA \cite{10.1145/3447755}, and POS \cite{wang2017algorithmic} as unsupervised methods, and adopt neural network-based supervised approaches including TSCAN \cite{liu2021multitasktemporalshiftattention}, PhysNet \cite{DBLP:confbmvcYuLZ19}, PhysFormer++~\cite{10.1007/s11263-023-01758-1}, and DeepPhys \cite{chen2018deepphys}. The \gls{hr} estimated from each modified video is then compared with the corresponding desired \gls{hr} $HR_d$ using standard error metrics, including Mean Absolute Error (MAE) and the Mean Absolute Percentage Error (MAPE).

\subsection{Qualitative Results}

Figure~\ref{fig:comparison} shows qualitative comparisons across multiple datasets. As can be observed from the original frames and the modified frames across multiple datasets, our method introduces minimal visual distortion, thereby preserving high visual fidelity in the reconstructed videos. The facial appearance and background remain consistent with the input, demonstrating that the proposed framework effectively preserves visual fidelity and edits \gls{hr}. For these experiments, we target three representative \gls{hr} values commonly encountered in practice: 60~bpm, 90~bpm, and 120~bpm. Moreover, from the zoomed-in regions shown in the figure, it can be observed that, due to the encoding–decoding pipeline of the \gls{3dvae} backbone, local distortions may occasionally appear in high-frequency areas such as edges and textures. These qualitative observations highlight the effectiveness of our approach in editing physiological signals without noticeably degrading perceptual quality.

\subsection{Quantitative Results}

\begin{table*}[t]
	\centering  
	\caption{Benchmark comparison between our method and PulseEdit~\cite{9680677} on video quality and \gls{rppg} estimation accuracy across multiple datasets. 
		PSNR and SSIM are reported for the output videos, while MAE and MAPE are computed for both input and output signals \textbf{relative to a reference \gls{hr} of 120 bpm} under the POS estimator. 
		Since the authors did not release their code base, did not release official code, we implement PulseEdit based on the descriptions provided in their work, adopting similar face detector, resizing, and temporal chunking procedures as in our pipeline for fairness. Our codebase will include implementation of PulseEdit.
		\textbf{Bold} denotes the best result.}
	\label{tab:benchmark}
	\begin{adjustbox}{max width=0.9\textwidth}
		\begin{tabular}{c c c c|cc|cc}
			\toprule    
			\multirow{2}{*}{\textbf{Dataset}} & \multirow{2}{*}{\textbf{Method}} 
			& \multirow{2}{*}{\textbf{PSNR(dB)} $\uparrow$} & \multirow{2}{*}{\textbf{SSIM} $\uparrow$} 
			& \multicolumn{2}{c|}{\textbf{Input \gls{rppg} Accuracy}} 
			& \multicolumn{2}{c}{\textbf{Output \gls{rppg} Accuracy}} \\
			\cmidrule(lr){5-6} \cmidrule(lr){7-8}
			& & & 
			& MAE(bpm) $\downarrow$ & MAPE(\%) $\downarrow$ 
			& MAE(bpm) $\downarrow$ & MAPE(\%) $\downarrow$ \\
			\midrule    
			
			\multirow{2}{*}{PURE~\cite{stricker2014video}}
			& Our Work     & 38.94 & \textbf{0.9761} & \multirow{2}{*}{50.63} & \multirow{2}{*}{42.20} & \textbf{10.36} & \textbf{11.34} \\
			& PulseEdit~\cite{9680677}    & \textbf{42.68} & 0.9720 &  &  & 16.71 & 12.26 \\
			\midrule
			
			\multirow{2}{*}{UBFC-rPPG~\cite{bobbia2017unsupervised}}
			& Our Work     & 40.04 & 0.9803 & \multirow{2}{*}{24.40} & \multirow{2}{*}{20.34} & \textbf{11.18} & \textbf{10.15} \\
			& PulseEdit~\cite{9680677}    & \textbf{43.08} & \textbf{0.9867} &  &  & 15.07 & 15.56 \\
			\midrule
			
			\multirow{2}{*}{MMPD~\cite{10340857}}
			& Our Work     & 37.87 & 0.9542 & \multirow{2}{*}{41.03} & \multirow{2}{*}{34.19} & \textbf{10.75} & \textbf{7.96} \\
			& PulseEdit~\cite{9680677}    & \textbf{41.72} & \textbf{0.9664} &  &  & 20.36 & 18.30 \\
			\bottomrule
		\end{tabular}
	\end{adjustbox}
\end{table*}

We consider a range of representative \gls{hr}s (60, 80, 100, and 120 bpm) as target values, and prompt our model to adjust the input video accordingly, modifying its \gls{hr} toward the specified target in the output.
Table~\ref{tab:metrics} presents a comparison of video quality and \gls{rppg} estimation accuracy between the input videos and the outputs produced by our framework across three benchmark datasets. 
For video quality, PSNR and SSIM are reported on the outputs relative to the inputs. Achieving around 40 dB PSNR and above 0.95 SSIM demonstrates that our method preserves high visual fidelity. 
For physiological signal analysis, we report MAE, and MAPE with respect to the target \gls{hr}. 
The input signals exhibit large deviations from the target. 
In contrast, the output signals consistently achieve markedly reduced MAE and MAPE, confirming the success of our \gls{hr} modification in aligning estimations with the desired target. 
We further observe that the outputs also reduce variability across subjects and trials, as indicated by smaller standard deviations, suggesting that our method not only corrects the mean error but also stabilizes estimation outcomes. In addition, we report in the supplementary material section 2 the best achievable visual quality with a plain 3D Causal VAE baseline without any modification layers, which serves as an upper bound for reconstruction fidelity under such a framework.
Furthermore, Table~\ref{tab:estimators} compares the results across different rPPG estimators that we trail on, where we use 120 bpm as the target \gls{hr} on PURE dataset. The results show that all unsupervised and supervised estimators are successfully misled by our framework into perceiving the manipulated signals as being close to 120 bpm. This highlights the robustness of our approach against diverse estimators.

\subsection{Benchmark Comparison}
\begin{table}[htbp]
	\centering
	\caption{MAE and MAPE result for input/output videos w.r.t. a fixed reference HR of 120 bpm, showing our method successfully fools a range of unsupervised and supervised rPPG estimators.}
	
	\label{tab:estimators}
	\begin{adjustbox}{width=0.48\textwidth}
		\begin{tabular}{c c|cc|cc}
			\toprule
			\multirow{2}{*}{\textbf{Dataset}} & \multirow{2}{*}{\makecell{\textbf{rPPG}\\\textbf{Estimator}}} 
			& \multicolumn{2}{c|}{\textbf{Input \gls{rppg} Accuracy}} 
			& \multicolumn{2}{c}{\textbf{Output \gls{rppg} Accuracy}} \\
			\cmidrule(lr){3-4} \cmidrule(lr){5-6}
			& & MAE(bpm) $\downarrow$ & MAPE(\%) $\downarrow$ 
			& MAE(bpm) $\downarrow$ & MAPE(\%) $\downarrow$ \\
			\midrule
			\multirow{7}{*}{\makecell{PURE\\\cite{stricker2014video}}} 
			& POS          & 50.63 & 42.20 & 10.36 & 11.34 \\
			& PCA          & 57.08 & 47.57 & 14.54 & 16.62 \\
			& CHROM        & 49.61 & 41.34 & 14.14 & 11.45 \\
			& TSCAN        & 42.23 & 35.19 & 13.68 & 12.03 \\
			& DeepPhys     & 49.37 & 41.14 & 11.75 & 11.84 \\
			& PhysNet      & 37.40 & 41.17 & 10.07 & 11.58 \\
			& PhysFormer++ & 38.95 & 32.46 & 10.29 & 10.53 \\
			\midrule
		\end{tabular}
	\end{adjustbox}
\end{table}         
We compare our framework against PulseEdit~\cite{9680677} across three benchmark datasets in terms of both video quality and \gls{rppg} estimation accuracy, as shown in Table~\ref{tab:benchmark} with a fixed target \gls{hr} of 120 bpm. Since the authors of PulseEdit did not release official code, we implemented their method following the description in their paper to the best of our ability. Overall, PulseEdit achieves slightly higher PSNR values, which can be attributed to its purely signal processing–based design. In contrast, our approach relies on a lossy encoding–decoding pipeline, which naturally leads to slightly lower PSNR values compared to purely signal processing–based methods. Due to the lossy compression inherent in the 3D VAE encoding-decoding process, our method achieves slightly lower PSNR values compared to direct signal processing approaches (see Appendix~\ref{sec:supp_vae_choice}). Both methods achieve comparable SSIM values, suggesting that our framework is able to preserve structural consistency of the video content despite the lossy compression. More importantly, our method consistently delivers superior performance on \gls{hr} modification. Specifically, the output signals produced by our framework yield lower MAE and MAPE with respect to the reference \gls{hr}, indicating that \gls{rppg} estimators are more effectively guided toward the desired target. In addition, compared to PulseEdit, our framework demonstrates stronger robustness across datasets with diverse characteristics: even for MMPD~\cite{10340857} and UBFC-rPPG~\cite{bobbia2017unsupervised}, which involve significant head movements or varying skin tones, our method maintains stable performance.
This highlights the trade-off between strict pixel-level fidelity and successful physiological signal editing.

\subsection{Ablation Study}

Table~\ref{tab:ablation} reports the ablation study evaluating the contribution of each component in our framework. We begin with a baseline configuration (Setting a), which employs a spatio-temporal layer setting as in Fig.~\ref{fig:settings}(a). This setting achieves limited performance since it does not account for the temporal coherence of the \gls{rppg} signal.
Incorporating temporal attention with \gls{adaln} improves both video quality and physiological consistency, as temporal modeling better preserves the temporal coherence of \gls{rppg} signals. Adding \gls{film}-based conditioning in the decoder further enhances performance by injecting the target condition into the generative process, resulting in improved fidelity. Finally, introducing curriculum-based frequency loss yields the best overall results, boosting PSNR/SSIM while reducing MAE/MAPE, demonstrating that progressive frequency loss effectively guides the network to first reconstruct visual content and then gradually align with the desired \gls{hr} pattern.

\begin{table}[htpb]
	\centering
	\footnotesize
	
	\renewcommand{\arraystretch}{1.2} 
	
	\caption{Ablation study of key components in our framework. MAE/MAPE for \gls{hr} estimation is conducted under POS estimator in PURE dataset~\cite{stricker2014video}. Each row shows the effect of adding a component to baseline (Setting a). \textbf{Bold} denotes the best result.}
	\label{tab:ablation}
	
	\begin{adjustbox}{width=0.48\textwidth}
		\begin{tabular}{lcccc}
			\hline
			\textbf{Configuration} & \textbf{PSNR} ↑ & \textbf{SSIM} ↑ & \textbf{MAE} ↓ & \textbf{MAPE} ↓ \\
			\hline
			Baseline Setting (a) & 31.84 & 0.8807 & 27.63 & 25.63 \\
			+ Temporal Attn. + \gls{adaln} & 34.82 & 0.9242 & 18.28 & 19.84 \\
			+ Decoder \gls{film} conditioning & 35.45 & 0.9153 & 14.92 & 13.91 \\
			+ Curriculum-based freq. loss & \textbf{38.94} & \textbf{0.9761} & \textbf{10.36} & \textbf{11.20} \\
			\hline
		\end{tabular}
	\end{adjustbox}
	
\end{table}

%% file: sec/5_discussion.tex
\section{Conclusion}

We presented \textit{PhysioLatent}, a framework for editing camera-based physiological signals in facial videos while preserving visual fidelity. By conditioning a pretrained \gls{3dvae} latent space with text-embedded \gls{hr} prompts and combining spatio-temporal attention with \gls{adaln} and \gls{film}-based conditioning, our method achieves controllable \gls{rppg} manipulation. On multiple datasets, \modelname attains high perceptual quality and low modulation error, and also supports an \gls{hr} removal mode that suppresses periodic components without harming appearance. Finally, its text-driven design integrates naturally with VLM- and diffusion-based pipelines, enabling both privacy-preserving editing and the creation of physiologically consistent synthetic datasets. 

%% file: sec/X_suppl.tex
\clearpage
\setcounter{page}{1}
\setcounter{section}{0}
\setcounter{figure}{0}
\setcounter{table}{0}
\maketitlesupplementary

\section{Discussion}
While our design opens up promising research opportunities and supports potential real-world applications, several limitations remain to be addressed. In the following, we discuss practical applications, outline future research potentials, and review the remaining challenges accordingly.

\subsection{Applications}
\label{sec:application}
\textbf{\gls{hr} Removal Mode.} Beyond targeted \gls{hr} editing, our framework can also operate in a removal model. In this setting, the objective is not to replace an existing \gls{hr} with a new target, but rather to eliminate the physiological trace. Technically, this can be achieved by modifying the loss functions such that both the frequency loss $\mathcal{L}_{\text{freq}}$ and wave loss $\mathcal{L}_{\text{wave}}$ are computed against pure Gaussian noise instead of a physiologically meaningful signal. At the same time, the textual conditioning prompt can be reformulated as \texttt{"Remove heart rate signal"}. In effect, the model is guided to suppress periodic \gls{rppg} components, producing videos in which \gls{hr} information no longer presents while preserving visual fidelity.Table~\ref{tab:remove} reports our results demonstrating the effectiveness of the proposed removal mode. High PSNR and SSIM show that our approach preserves visual quality after signal suppression. 
For the physiological signals, we report both the input SNR and output SNR, along with the change in  ($\Delta$SNR), defined as the difference between the spectral power around the ground-truth \gls{hr} frequency and the remaining spectrum before and after removal. 
The consistently negative $\Delta$SNR values, indicate a marked suppression of periodic \gls{rppg} components, validating the effectiveness of our removal strategy.

\begin{table}[h]
	\centering
	\footnotesize
	\caption{Quantitative results in the \textbf{\gls{hr} removal mode}. 
		PSNR and SSIM measure the visual fidelity of the output videos. 
		Input SNR and output SNR denote the SNR of the original and modified rPPG signals, respectively. 
		$\Delta$SNR represents their difference, where more negative values indicate stronger suppression of the \gls{hr} component.}
	\label{tab:remove}
	
	\renewcommand{\arraystretch}{1.3} 
	
	\begin{adjustbox}{max width=\textwidth}
		\begin{tabular}{lccccc}
			\hline
			\textbf{Dataset} & \textbf{PSNR} ↑ & \textbf{SSIM} ↑ & \makecell{\textbf{Input}\\\textbf{SNR}} & \makecell{\textbf{Output}\\\textbf{SNR}} & \textbf{$\Delta$SNR}\\
			\hline
			PURE~\cite{stricker2014video}            & 37.28 & 0.9532 & 26.82 & 16.37 & \textbf{-10.45} \\
			UBFC-rPPG~\cite{bobbia2017unsupervised}  & 38.53 & 0.9681 & 24.94 & 15.02 & \textbf{-9.92}  \\
			MMPD~\cite{10340857}                     & 36.80 & 0.9447 & 27.35 & 16.05 & \textbf{-11.31} \\
			\hline
		\end{tabular}
	\end{adjustbox}
\end{table}

\begin{figure}[h]
	\centering
	\includegraphics[width=\linewidth]{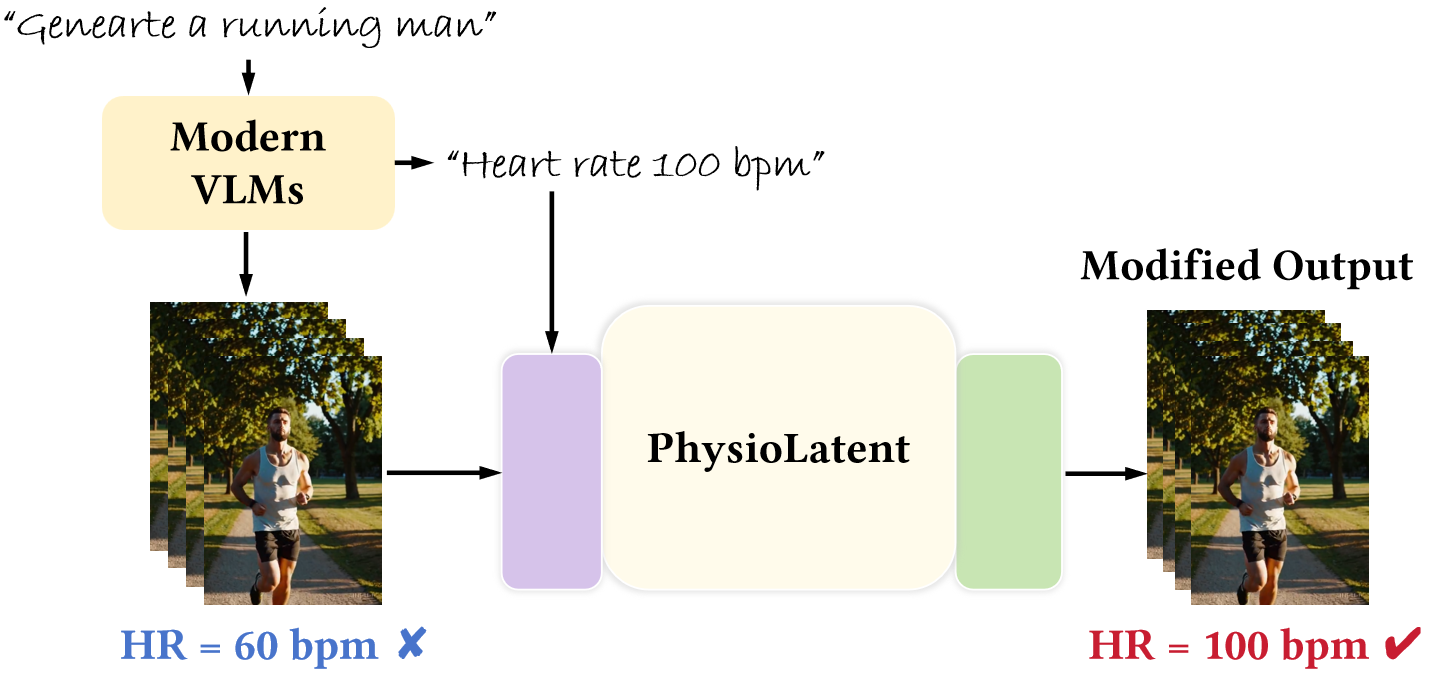}
	\caption{Illustration of a text-driven use case. A VLM generates both a video (e.g., "running man") and an associated \gls{hr} prompt (e.g., \texttt{"Heart rate 100 bpm"}). 
		The raw generated video may not exhibit a physiologically consistent \gls{hr}, whereas our model \textbf{\modelname}adjusts the \gls{rppg} signal accordingly, producing a video with both semantically correct content and physiologically accurate dynamics.}
	\label{fig:application}
\end{figure}

\textbf{\gls{hr} in generated videos.} One practical use case of our framework is that, our selected 3D VAE operates in the same latent space as modern generative backbones, potentially enabling direct integration with foundation-model pipelines. Prior studies have shown that videos generated by generative models often exhibit unrealistic \gls{hr} patterns~\cite{9304909,9141516}. To address this limitation, a Visual Language Model (VLM) can be employed to automatically determine an appropriate \gls{hr} while generating the corresponding video clip in response to contextual queries (e.g., "Generate a resting scenario with a calm subject" or "Generate a running man"). The chosen \gls{hr} value can then be expressed in natural language form—such as \texttt{"Heart rate 100 bpm"}—and provided as a conditioning prompt to our model. In this way, the generated videos not only reflect the intended semantic content but also exhibit physiologically consistent \gls{hr} patterns, thereby supporting the creation of more realistic and precise synthetic datasets for downstream tasks, as shown in Fig.~\ref{fig:application}.  It is also worth noting that our selected \gls{3dvae} operates in the same latent space as modern generative backbones, ensuring direct compatibility with foundation-model pipelines. This design choice allows VLM-driven prompts and latent manipulations to interface seamlessly with our framework, enabling controllable and physiologically consistent video editing in future generative systems. While our current use of CLIP may appear similar to a categorical index, its formulation anticipates such extensions, positioning our method as a bridge between semantic control and physiological realism.

\subsection{Limitations and Future Work}

\textbf{Lossy video reconstruction.}  
Our approach operates in the latent space of a \gls{3dvae}, which inherently involves a lossy process. Even with the strongest available \gls{3dvae} backbone, reconstruction fidelity is not perfect, and local distortions may appear in the output videos, especially in high-frequency regions, i.e., regions with edges or textures (as shown in Fig.~\ref{fig:comparison}), leading to a slight drop in PSNR compared to signal-processing–based baselines. One promising direction is to replace the \gls{3dvae} backbone with emerging representations such as Gaussian Splatting~\cite{kerbl20233dgaussiansplattingrealtime}, which have recently been explored for video embeddings~\cite{wang2025gsvcefficientvideorepresentation}. Such variants could enable rPPG editing directly in the Gaussian parameter space, potentially improving reconstruction fidelity while preserving subtle physiological signals.  

\textbf{Lack of \gls{hrv} modeling.}  
Our current design does not explicitly model \gls{hrv}. The employed wave loss enforces alignment with a simple sinusoidal template at the target frequency, which facilitates \gls{hr} manipulation but cannot capture the richer temporal dynamics associated with \gls{hrv}.  
A natural extension would be to incorporate temporal generative priors such as state-space models~\cite{ROSAS2024107857} or neural ODEs~\cite{nazaret2023modeling}, which can represent richer stochastic dynamics, thereby enabling explicit editing of both mean \gls{hr} and its variability.  

\textbf{Dataset generalization.}  
Our evaluation is currently limited to benchmark datasets, while generalization to in-the-wild conditions with challenging illumination or motion remains unexplored. Expanding the training set to more diverse scenarios would improve robustness. Large-scale pretraining with synthetic-to-real domain adaptation, or leveraging multimodal foundation models that align video with physiological signals, could significantly enhance generalization in unconstrained environments.  

\textbf{Incomplete frequency suppression.}  
In some cases, the suppression of the original \gls{hr} component may be incomplete: while the PSD peak shifts toward the target frequency, residual energy at the original frequency can persist, yielding multi-peak spectra and potentially confusing downstream estimators. Frequency-domain consistency losses may help achieve better suppression, ensuring that residuals are eliminated while preserving visual fidelity.  

\textbf{Prompt interval limitation.}  
During training, the prompts were randomly sampled within a fixed range of 60--120~bpm at 10~bpm intervals. As a result, when unseen intermediate \gls{hr} (e.g., 75~bpm) are provided as prompts, the model may fail to generate accurate outputs. Our Appendix.B includes examples using target \gls{hr}s of 75 bpm, demonstrating the model's interpolation behavior between the discrete prompt values, but also indicating the limitations to be tackled in our future work. A more robust approach would be continuous prompt conditioning, where \gls{hr}s are drawn from a continuous distribution rather than discrete intervals. We also provide several futuristic solutions to this issue in Appendix~\ref{sec:sec_app_b}.

\textbf{Potential misuse and ethical considerations.}  
While the ability to remove or alter \gls{hr} signals expands the flexibility of our framework, it also introduces the possibility of misuse. In particular, fabricated or misleading \gls{hr} prompts could be presented as genuine to manipulate perceived health conditions. Such misuse might enable malicious actors to falsify physiological evidence, for instance in the context of insurance claims, medical record falsification, or biometric-based authentication. To mitigate misuse, technical safeguards such as watermarking physiological edits or cryptographic verification of unaltered signals could be developed. Alongside, ethical guidelines and responsible usage policies will be crucial for ensuring that these tools are deployed safely in medical and biometric applications. 

\textbf{Limited input prompt.} 
Our current framework has been validated only using a predefined set of prompts, which ensures controlled evaluation but limits the diversity of scenarios. Extending to a broader and more flexible prompt space remains an important direction for future work, particularly in the context of generative models. For example, when integrated into a text-to-video pipeline, one may provide a single composite prompt that specifies both semantic content and physiological cues (e.g., \texttt{"Generate a running man with heart rate 120 bpm"}). Realizing such unified conditioning would require handling more natural and diverse linguistic expressions, thereby enabling tighter integration between semantic intent and physiologically consistent video synthesis—a promising future direction for our work.
\section{VAE Backbone Selection and Analysis}
\label{sec:supp_vae_choice}
To ensure that our framework is supported by the most reliable backbone, we perform a systematic comparison across several candidate \gls{3dvae} models. The objective of this analysis is to demonstrate that we have carefully selected the best-performing VAE available to us. We compare 3D Causal VAE~\cite{yang2024cogvideox}, TAESDV (\url{https://github.com/madebyollin/taesd}), and two variants of Video-VAE~\cite{xing2025videovae}. Each candidate is evaluated under a plain encoding–decoding pipeline, without any additional modification layers, so that the intrinsic reconstruction quality of the backbone can be fairly assessed. 
\begin{table}[h]
	\centering
	\footnotesize
	\caption{Comparison of different \gls{3dvae} backbones on the PURE dataset under plain encoding–decoding (without modification layers). The 3D Causal VAE achieves the best trade-off and is therefore adopted in our framework.}
	\label{tab:vae_comparison}
	\begin{tabular}{lcc}
		\toprule
		\textbf{3D VAE Backbone} & \textbf{PSNR$\uparrow$} & \textbf{SSIM}$\uparrow$ \\
		\midrule
		\textbf{3D Causal VAE} & \textbf{36.90} & \textbf{0.9621} \\
		TAESDV & 22.35 & 0.7637 \\
		Video-VAE (4 channels) & 35.38 & 0.9319 \\
		Video-VAE (16 channels) & 35.68 & 0.9527 \\
		\bottomrule
	\end{tabular}
\end{table}

Table~\ref{tab:vae_comparison} reports the results on the PURE dataset, measuring the reconstruction quality of videos after passing through different \gls{3dvae} backbones. Based on this comparison, we adopt the 3D Causal VAE as the backbone for our framework, since it consistently achieves superior reconstruction fidelity while maintaining robustness across different inputs.

\section{Discussion on Prompt Sampling Strategy}
\label{sec:sec_app_b}
\begin{figure}[h]
	\centering
	\includegraphics[width=\linewidth]{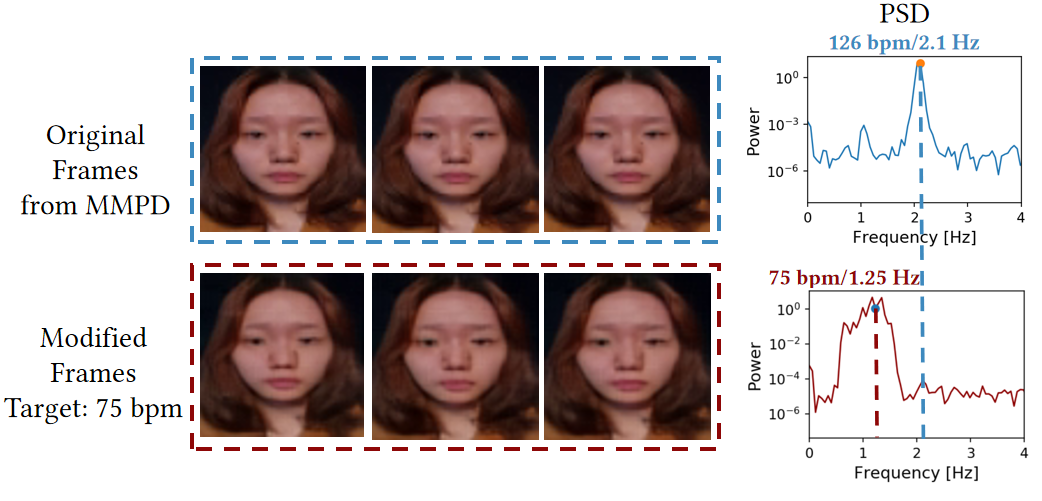}
	\caption{Qualitative example with an unseen prompt of 75~bpm. While the PSD peak is partially shifted toward 75~bpm, residual energy at neighboring frequencies remains, demonstrating limited interpolation capability. Source images are from MMPD~\cite{10340857}.}
	\label{fig:prompt75}
\end{figure}

During training, the \gls{hr} prompts were randomly sampled within a fixed range of 60--120~bpm at 10~bpm intervals. While this strategy ensures sufficient coverage of typical \gls{hr} values, it inherently restricts the model to a discrete set of conditions. As a result, when the model encounters unseen intermediate prompts (e.g., 75~bpm), the generated outputs may not align precisely with the desired target frequency. This limitation arises because the model has not been explicitly exposed to such in-between cases during training, thereby reducing its interpolation capability.  \\
To illustrate this limitation, we provide an example where the input prompt is set to 75~bpm. As shown in Fig.~\ref{fig:prompt75}, although the PSD of the generated signal exhibits a peak shift toward the intended 75~bpm frequency, the alignment is imperfect. Residual energy remains at the nearest seen training frequencies (70~bpm and 80~bpm), leading to a multi-peak spectrum. This observation highlights the interpolation gap induced by the discrete prompt sampling strategy.  

We identify two potential directions to address this limitation in future work.  
First, adopting a continuous conditioning strategy, where target values are drawn from a continuous random distribution over the entire range rather than from a small set of fixed discrete intervals (e.g., 60, 70, 80), could improve generalization to unseen values. Second, introducing data augmentation in the frequency domain, such as randomized frequency shifts or adversarial perturbations, may further enhance the robustness of the learned conditioning mechanism.

\section{In-the-wild Results}
\label{sec:appendix_wild}
\begin{figure}[h]
	\centering
	\includegraphics[width=\linewidth]{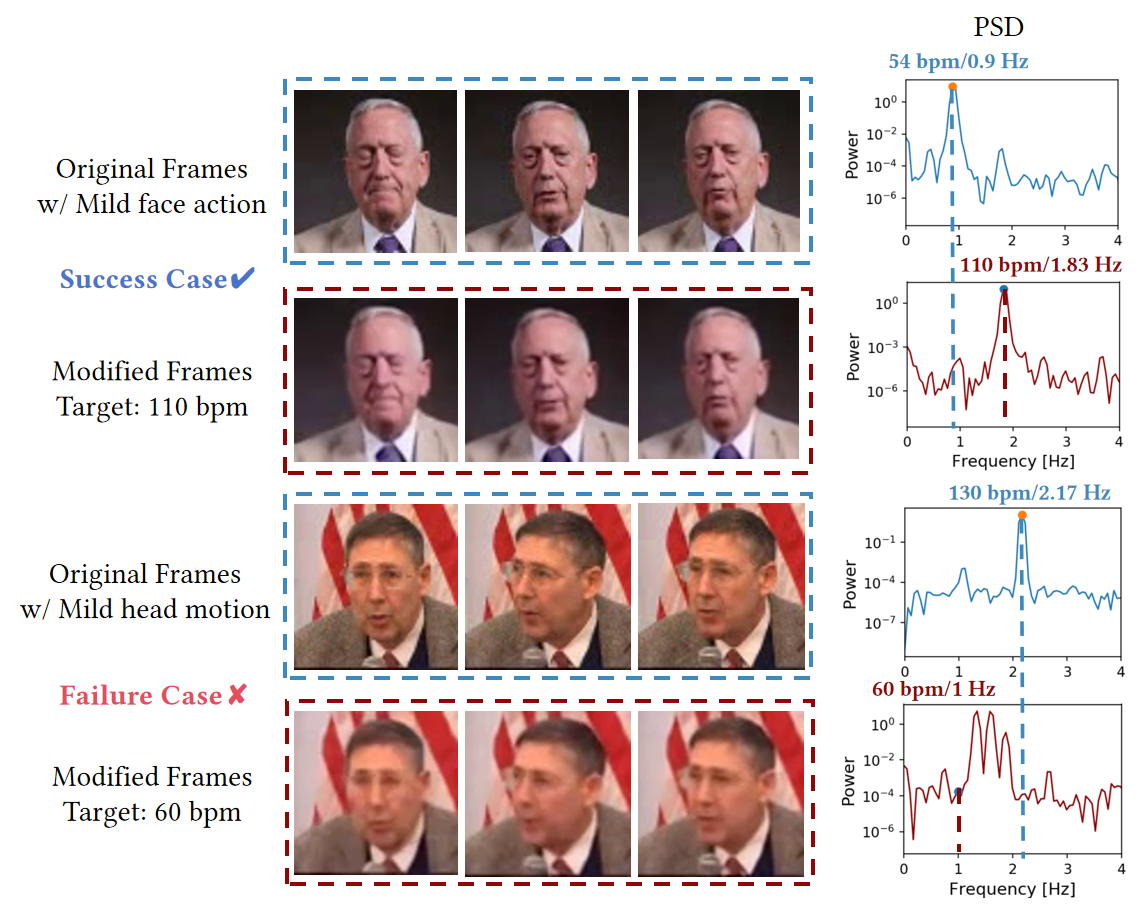}
	\caption{Representative in-the-wild results from the TalkingHead1KH dataset~\cite{wang2021facevid2vid}. 
		Our method performs well on videos with pronounced facial actions, achieving accurate and visually consistent \gls{hr} modulation, but struggles with large head-pose variations, which may lead to less precise \gls{hr} modification.}
	\label{fig:wild_cases}
\end{figure}

To complement our benchmark experiments, we also evaluated \modelname on a small set of in-the-wild videos with challenging illumination and motion conditions from the TalkingHead1KH dataset~\cite{wang2021facevid2vid}. 
Figure~\ref{fig:wild_cases} presents representative examples, including both successful and unsuccessful cases. 
Our method performs well on videos with pronounced facial actions, where heart-rate modulation remains accurate and visually consistent. 
However, for sequences with large head-pose changes, the model struggles to maintain temporal coherence, leading to less precise heart-rate modification. Also, since the input videos are in-the-wild, they are out of the distribution of the training set, affecting visual quality.

